\documentclass[review]{elsarticle}

\usepackage{lineno,hyperref}
\modulolinenumbers[5]

\journal{Journal of \LaTeX\ Templates}

\makeatletter
\def\ps@pprintTitle
{%
   \let\@oddhead\@empty
   \let\@evenhead\@empty
    \def\@oddfoot{Preprint. Work in progress.\hfil}
   \let\@evenfoot\@oddfoot
}
\makeatother
\usepackage{times}
\usepackage{epsfig}
\usepackage{graphicx}
\usepackage{amsmath}
\usepackage{amssymb}

\usepackage{booktabs}       % professional-quality tables
\usepackage{multirow}
\usepackage{adjustbox}
\usepackage{bm}
\newcommand{\ie}{\textit{i}.\textit{e}.}
\newcommand{\eg}{\textit{e}.\textit{g}.}

\newcommand{\etal}{{\it et al.}}
\usepackage[parfill]{parskip}

\usepackage[usenames]{color}
\definecolor{blue}{rgb}{0,0,1}
\newcommand{\blue}[1]{\textcolor{blue}{#1}}

\definecolor{red}{rgb}{1,0,0}

\definecolor{black}{rgb}{0,0,0}
\newcommand{\black}[1]{\textcolor{black}{#1}}
\usepackage{mathtools}
\DeclarePairedDelimiterX{\infdivx}[2]{(}{)}{%
  #1\;\delimsize\|\;#2%
}
\newcommand{\infdiv}{\infdivx}
\usepackage{multirow}

\begin{document}

\begin{frontmatter}

\title{FoCL: Feature-Oriented Continual Learning for \\ Generative Models}
% \tnotetext[mytitlenote]{Preprint. Work in progress.}

\author[imagia,concordia]{Qicheng Lao\corref{mycorrespondingauthor}}
\ead{qicheng.lao@gmail.com}

% \address{Radarweg 29, Amsterdam}
% \fntext[myfootnote]{Since 1880.}

% \author[mymainaddress,mysecondaryaddress]{Elsevier Inc}

\author[concordia]{Mehrzad Mortazavi}
\author[mcgill]{Marzieh Tahaei}
\author[imagia]{Francis Dutil}
\author[concordia]{Thomas Fevens}
\author[imagia]{Mohammad Havaei\corref{mycorrespondingauthor}}
\ead{mohammad@imagia.com}

\address[imagia]{Imagia Inc., Montreal, Canada}
\address[concordia]{Department of Computer Science and Software Engineering, Concordia University, Montreal, Canada}
\address[mcgill]{Department of Biomedical Engineering, McGill University, Montreal, Canada}

\cortext[mycorrespondingauthor]{Corresponding authors}

\begin{abstract}
In this paper, we propose a general framework in continual learning for generative models: Feature-oriented Continual Learning (FoCL). Unlike previous works that aim to solve the catastrophic forgetting problem by introducing regularization in the parameter space or image space, FoCL imposes regularization in the feature space. We show in our experiments that FoCL has faster adaptation to distributional changes in sequentially arriving tasks, and achieves the state-of-the-art performance for generative models in task incremental learning. We discuss choices of combined regularization spaces towards different use case scenarios for boosted performance, \eg, tasks that have high variability in the background. Finally, we introduce a forgetfulness measure that fairly evaluates the degree to which a model suffers from forgetting. Interestingly, the analysis of our proposed \textit{forgetfulness score} also implies that FoCL tends to have a mitigated forgetting for future tasks.
\end{abstract}


\end{frontmatter}

% \linenumbers

\section{Introduction}
Generative models have shown great potential for generating natural images in {\it stationary} environments under the assumption that training examples are available throughout training, and are independent and identically distributed~({\it i.i.d.}). Continual Learning (CL), however, is the ability to learn from a continuous stream of data in a {\it non-stationary} environment, which entails that not only the distribution of the data is subject to change but also the nature of the task itself.
To learn under such circumstances, the model needs to have the {\it plasticity} to acquire new knowledge and {\it elasticity} to deal with catastrophic forgetting~\cite{farquhar2018towards}. 
However, neural networks have been shown to deteriorate severely on previous tasks when trained in a sequential manner~\cite{mccloskey1989}.

\begin{figure*}[!t]
	\centering
	\includegraphics[width=\textwidth]{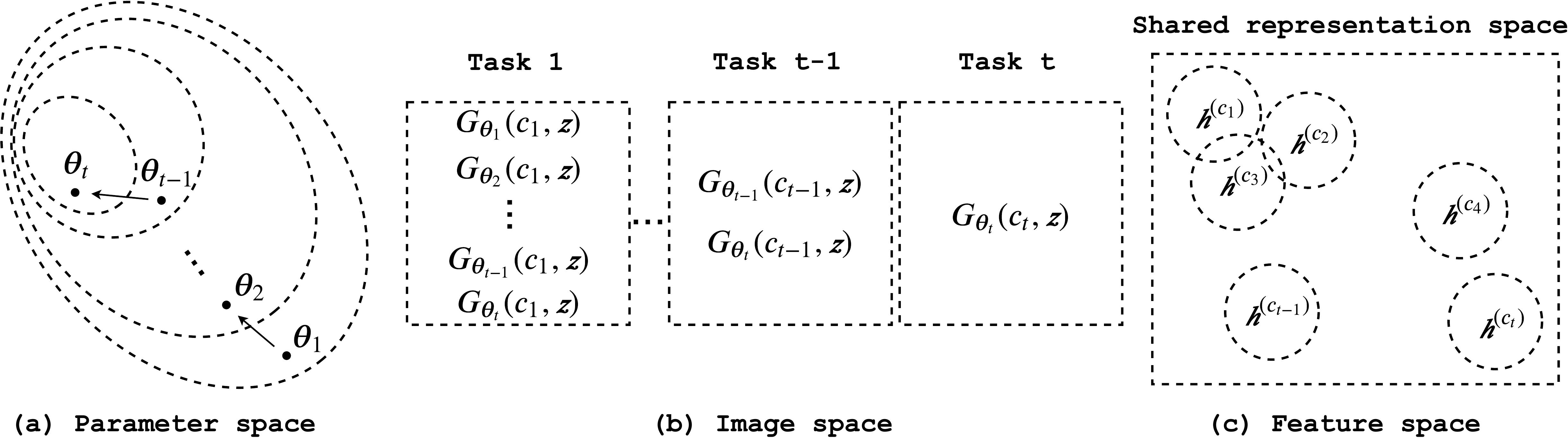}
	\caption{Approaches in continual learning from the perspective of regularization space. $G_{\bm{\theta}_t}(c_i, \bm{z})$ denotes for the images generated by the model at task $t$ for a previous task $i$~($c_i$ is the conditioning factor), and $\bm{h}^{(i)}$ denotes for the features representing task $i$.}
	\label{fig_spaces}
\end{figure*}

Current scalable~\footnote{Therefore other categories of approaches, such as memory rehearsal/replay and dynamic architectures, are not included here.} approaches to mitigate catastrophic forgetting can be grouped into two main categories: the {\it prior-based} approaches, where the model trained on previous tasks acts as a prior for the model training on the current task via a regularization term applied on the parameters \cite{kirkpatrick2017overcoming,zenke2017continual,nguyen2017variational}; and the {\it replay-based} approaches, where synthetic images are generated through a snapshot model to mimic the real data seen in previous tasks. These images are then used as additional training data to constrain the model from forgetting~\cite{shin2017continual,wu2018memory}.
Both categories deal with forgetting by different means of regularization: the prior-based approaches regularize in the parameter space, \ie, the parameters of the current model are regularized to stay close to that of previous model (illustrated in Figure~\ref{fig_spaces} (a)), whereas the replay-based approaches remember previous tasks by regularizing in the image space (Figure~\ref{fig_spaces} (b)). While both means of regularization have shown great success in the literature, each has its own limitations. For example, the strong regularization in the parameter space can limit the expressive power of the model in acquiring new knowledge, and the regularization in the image space on the other hand, may include the mapping of some undesired noisy information, since the image space is not fully representative of the underlying nuances of a task. 

In this work, we present a new framework in continual learning for generative models: Feature-oriented Continual Learning (FoCL), where we propose to address the catastrophic forgetting problem by imposing the regularization in the shared feature space (Figure~\ref{fig_spaces} (c)). Instead of the strong regularization in the parameter space or image space as done in previous works, we regularize the model in a more meaningful and therefore more efficient way in the feature representation space. %Intuitively, by doing so, the model should first be able to better preserve the underlying semantics of a task and second learn task agnostic representations which could be transferred into future tasks. For example, as shown in Figure~\ref{fig_1st}, FoCL captures the two backgrounds that are shareable across different tasks in the SVHN dataset, and in CIFAR10 and LSUN datasets, it learns the semantic structures across tasks, \ie, animals vs vehicles and indoor scenes vs outdoor scenes. We believe that learning such shareable features across different tasks could facilitate the remembering and therefore result in better performance.
Regularizing in the feature space is also related to regularization in the functional space introduced in a parallel work by Titsias~\etal~\cite{titsias2019functional}. In their approach, regularization is applied to distribution of task specific functions parameterized by neural networks, whereas in this work we regularize the high level features directly. Also their model is applied to supervised learning tasks but not generative tasks. 

We show in our experiments that our framework has faster adaptation to changes in the task distribution, and improves the performance of the generative models on several benchmark datasets in task incremental learning settings. To construct the feature space, several ways are proposed in this work: (1) an adversarially trained encoder, (2) distilled knowledge from the current model, and (3) a pretrained model that can represent our prior knowledge on the task descriptive information, which later on could be potentially applied to zero-shot learning scenarios. Moreover, we show that in some use case scenarios, FoCL performance can be further boosted by leveraging additional regularization in the image space, especially when the variability in the background is high that could heavily disturb the feature representation learning. Therefore, for those tasks, we propose to extend FoCL with combined regularization strategies. Finally, we introduce a new metric, \textit{forgetfulness score}, to fairly evaluate to which degree a model suffers from catastrophic forgetting. The analysis of the forgetfulness scores can offer complimentary information from a new perspective compared to current existing metrics, and it can also predict the potential performance on future tasks.

\section{Related work}
\subsection{Continual learning for generative models}
The two major approaches to generative models, namely generative adversarial networks (GANs)~\cite{goodfellow2014generative} and variational auto-encoders (VAEs)~\cite{kingma2013auto}, have been used in a continual learning setup, where the objective is to learn a model not only capable of generating examples of the current task but also previous tasks. To facilitate elasticity and preserve plasticity, one or a mixture of aforementioned solutions to catastrophic forgetting have been applied. The performance comparison of different generative models in the context of continual learning has also been investigated here~\cite{lesort2019generative}.

Nguyen \etal~\cite{nguyen2017variational} proposed variational continual learning (VCL), a VAE based model with separate heads (as conditioning factor) for different tasks. The task specific parameters help the plasticity of the model when dealing with different tasks. To address catastrophic forgetting, VCL uses variational inference within a Bayesian neural network to regularize the weights. In their framework, the posterior at the end of each task is set to be the prior for the next one. They also showed that rehearsal on a corpus of real examples from previous tasks can improve performance. 

Other works have taken GAN based approaches~\cite{seff2017continual,shin2017continual,wu2018memory}. Seff \etal~\cite{seff2017continual} used elastic weight consolidation (EWC~\footnote{EWC was first proposed by Kirkpatrick \etal~\cite{kirkpatrick2017overcoming} in supervised and reinforcement learning context.}) to prevent critical parameters for all previous tasks from changing. Shin \etal~\cite{shin2017continual} introduced deep generative replay (DGR) where a snapshot of the model trained on previous tasks is used to generate synthetic training data from previous tasks. The synthetic data is used to augment the training examples of the current task, thus providing a more \textit{i.i.d.} setup. Wu \etal~\cite{wu2018memory} took a similar approach to \cite{shin2017continual} but used a pixel-wise $l_2$ norm to regularize the model through the memory replay on previous tasks. 

While FoCL takes inspiration from \cite{shin2017continual} and \cite{wu2018memory}, in both of these approaches, the memory replay is performed to regularize the model in the image space. Whereas in FoCL, the memory replay is applied on the more task-representative feature space. 

\subsection{Feature matching for generative models}
Feature matching has been explored as a means for perceptual similarity in the context of generative adversarial networks~\cite{li2015generative,dosovitskiy2016generating,salimans2016improved,warde2016improving,nguyen2017plug}. Dosovitskiy \etal~\cite{dosovitskiy2016generating} used high-level features from a deep neural network to match visual similarity between real and fake images. They argued that while feature matching alone is not enough to provide a good loss function, it can lead to improved performance if used together with an objective on the image. Salimans \etal~\cite{salimans2016improved} showed that feature matching as an auxiliary loss function can lead to more stable training and improved performance in semi-supervised scenarios.  

Contrary to previous works that use feature matching to align the model distribution to the empirical data distribution, %. On the contrary 
we use feature matching as a regularizer to remember knowledge from previous tasks. % the model has learned.
Intuitively, matching in the feature space allows us to go beyond pixel-level information to focus more on remembering factors of variation contributing to a particular task, and those that are shareable among different tasks.

\section{Framework}
\subsection{Problem formalisation}
Given a stream of generative tasks $t \in [1, 2, ..., T]$ arrived sequentially, each of which has its own designated dataset $\bm{x}^{(t)}$ %$S_t = (\bm{X}^{(t)}, \bm{y}^{(t)})$
\footnote{For consistency, we use superscript throughout this paper to denote the task that the data or variable belongs to.}, the goal of continual learning in generative models is to learn a unified model parameterized by $\bm{\theta}$, such that $p_{\bm{\theta}}(\bm{x}^{(t)}) = p_{data}(\bm{x}^{(t)})$ for all $t \in [1, 2, ..., T]$. In a conditional generator setting, $p_{\bm{\theta}}(\bm{x}^{(t)}) = p_{\bm{\theta}}(\bm{x}|c_t)$, where $c_t$ is the category label for task $t$. The challenge comes from the \textit{non-i.i.d.} assumption in continual learning that for the current task $t$, the model has no access to the real data distributions for previous tasks. Therefore, it is critical that the model retains in memory all the knowledge learned in previous tasks (\ie, $p_{\bm{\theta}_{t-1}}$) while solving the current new task. This can be formulated as the following: 
\begin{equation}
p_{\bm{\theta}_t}(\bm{x} | c_i) =\left\{
  \begin{array}{@{}ll@{}}
    p_{data}(\bm{x}^{(i)}) & \quad i = t \\
    p_{\bm{\theta}_{t-1}}(\bm{x} | c_i) &  1 \leq i < t,
%   p_{data}(S^{(i)}) & i = t,
  \end{array}\right.
  \label{eq:general_objective}
\end{equation}
given the assumption that $p_{\bm{\theta}_{t-1}}(\bm{x} | c_i)$ estimates the data distributions of previous tasks incrementally well for all $ i < t$. As described earlier, the current continual learning approaches for solving eq.~(\ref{eq:general_objective}) mostly fall into two main categories depending on the regularization space: parameter space~\cite{kirkpatrick2017overcoming,zenke2017continual,nguyen2017variational} and image space~\cite{shin2017continual,wu2018memory}.

\subsection{Feature oriented continual learning}
Here, we propose a new framework in continual learning for generative models, where we focus the regularization in the feature space, by introducing an encoder function $f$ that maps images $\bm{x}$ into low-dimensional feature representations $\bm{h} = f(\bm{x})$. To mitigate catastrophic forgetting, we regularize the model to remember previous tasks by explicitly matching the high-level features learned through previous tasks. %that represent different tasks. %task representative high-level features
This updates the problem defined in eq.~(\ref{eq:general_objective}) to:
\begin{equation} \label{eq:general_objective2}
\begin{split}
p_{\bm{\theta}_t}(\bm{x} | c_i) &= p_{data}(\bm{x}^{(i)}) \quad\quad i = t \\
p_{\bm{\theta}_t}(\bm{h} | c_i) &=  p_{\bm{\theta}_{t-1}}(\bm{h} | c_i) \quad 1 \leq i < t.
\end{split}
\end{equation}
% $p_{\bm{\theta}_t}(\bm{x} | c_i)$ and $p_{\bm{\theta}_{t-1}}(\bm{x} | c_i)$ for all $i < t$. %In practice, this is usually done through alignment of examples generated by the two generative processes.

Intuitively, the change of regularization to feature space has two potential benefits. First, it allows the model to focus on matching more representative information instead of parameters or pixel-level information, \ie, not all pixels in the image space contribute equally for each task. Second, in cases where task representative information (\ie, task descriptors) is available, we could leverage that information directly in the alignment process without solely relying on $p_{\bm{\theta}_{t-1}}$ from the previous task. In fact, the matching in the feature space can also be viewed as a generalized version of the matching in the image space, which allows more flexibility in the regularization during optimization. %, \ie, the matching of images guarantees the matching of features, but not vice versa. Matching in the image space can be recovered if we learn an identity encoder function $f(x)=x$.

To maintain generality, we denote $D\infdiv{p}{q}$ as any divergence function that measures the disparity between distributions $p$ and $q$. Therefore, we can write the objective function for solving eq.~(\ref{eq:general_objective2}) as:
\begin{equation}
\begin{split}
  \mathcal{L}_{\text{FoCL}}(\bm{\theta}_t) 
  &= \blue{\underbrace{\black{D\infdiv{p_{data}(\bm{x}^{(t)})}{p_{\bm{\theta}_t}(\bm{x} | c_t)} }}_{\text{current task}}} + \blue{\underbrace{\black{ \sum_{i=1}^{t-1}\lambda_t D'\infdiv{p_{\bm{\theta}_{t-1}}(\bm{h}|c_i)}{p_{\bm{\theta}_t}(\bm{h}|c_i)}}}_{\parbox{62pt}{\scriptsize\centering previous tasks}}},
  \label{eq:feature_objective}
\end{split}
\end{equation}
where $D$ and $D'$ are appropriate instances of divergence functions. Common choices for the divergence function can include Kullback-Leibler divergence $D_{KL}\infdiv{p}{q}$, Jensen-Shannon divergence $D_{JS}\infdiv{p}{q}$ and Bregman divergence $D_{\phi}\infdiv{p}{q}$, where $\phi$ is a continuously-differentiable and strictly convex function. 
% \paragraph{VAE setup} 
For instance, in a GAN setup, minimizing the first divergence term in eq.~(\ref{eq:feature_objective}) (\textit{current task}) can be expressed as a minmax game: 
\begin{equation} \label{eq:wgan_objective}
    \min_{\bm{\theta}_t} \max_{\bm{\omega}} 
    \mathbb{E}_{{\bm{x}} \sim p_{\bm{\theta}_t}(\bm{x}|c_t)}[\psi_{\bm{\omega}}({\bm{x}})] -
    \mathbb{E}_{\bm{x} \sim p_{data}(\bm{x}^{(t)})}[\psi'_{\bm{\omega}}(\bm{x})],
\end{equation}
where $\psi_{\bm{\omega}}$ and $\psi'_{\bm{\omega}}$ are appropriate functions parameterized by $\bm{\omega}$ based on the chosen divergence function~\cite{grover2018flow}.
In a VAE setup, minimizing $D_{KL}\infdiv{p_{data}(\bm{x}^{(t)})}{p_{\bm{\theta}_t}(\bm{x} | c_t)}$ %between the data distribution and model distribution 
is equivalent to maximizing the marginal log-likelihood over data:
\begin{equation}
\begin{split}
    \max_{\bm{\theta}_t} \mathbb{E}_{\bm{x} \sim p_{data}(\bm{x}^{(t)})}[\log p_{\bm{\theta}_t}(\bm{x}|c_t)],
    \label{eq:vae_objective}
\end{split}
\end{equation}
which is then often approximated by the evidence lower bound (ELBO)~\cite{kingma2013auto}.
For the second divergence term in eq.~(\ref{eq:feature_objective}) (\textit{previous tasks}), the matching of empirical feature distributions could be done through the alignment of feature examples generated by the two generative processes ($p_{\bm{\theta}_t}$ and $p_{\bm{\theta}_{t-1}}$). 

As a proof of concept, in this work, we choose to use Wasserstein distance $W(p, q) = \underset{\gamma \in \Pi(p,q)}{\text{inf}} \mathbb{E}_{(x,y) \sim \gamma}[\left\|x-y\right\|]$ 
as the distance function for $D$ (\ie, in a Wasserstein GAN setup~\cite{arjovsky2017wasserstein}), and for the feature matching divergence $D'$, we experiment with both Bregman divergence with $\phi(x) = \left\|x\right\|^2$ (\ie, feature-wise $l_2$ loss) (Section~\ref{sec_prior_knowledge}) and Wasserstein distance (Section~\ref{exp_end_to_end}) in this study. 

\begin{figure*}[t]
	\centering
	\includegraphics[width=\textwidth]{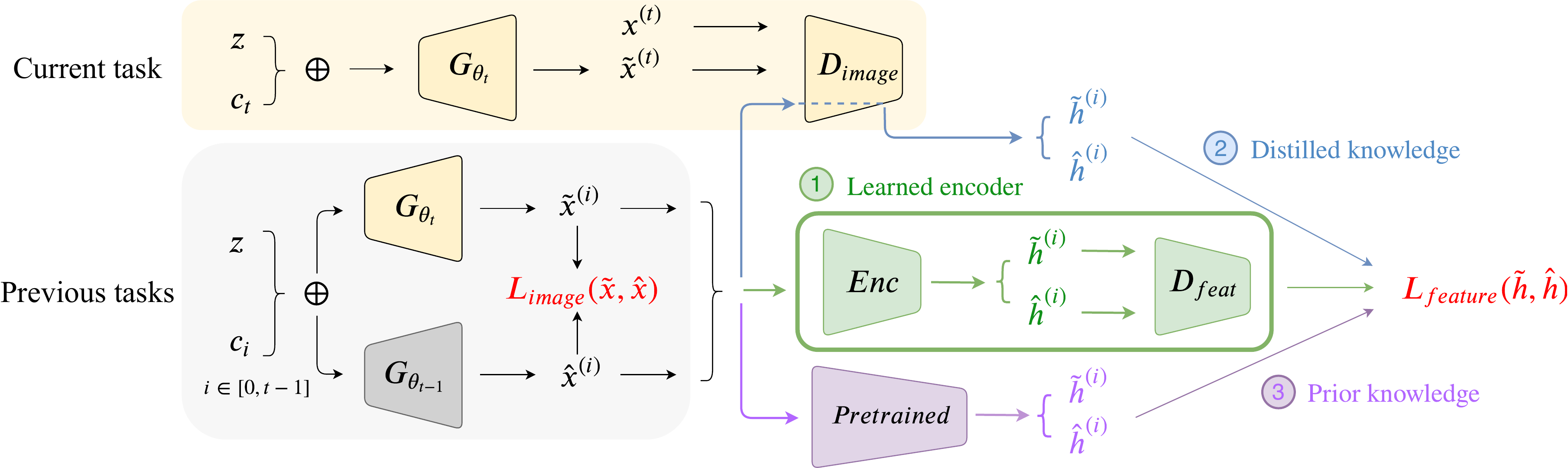}
	\caption{We propose three alternatives for constructing the feature space: (1) Adversarially learned encoder, (2) Distilled knowledge from the image discriminator, and (3) Prior knowledge from a pretrained model. \textit{Top}: learning current task, \textit{bottom}: replay from memory for previous tasks.}
	\label{fig_model}
\end{figure*}

\subsection{Constructing the feature space}
As illustrated in Figure~\ref{fig_model}, we present three alternatives to construct the feature space. For simplicity of notation, we denote $\Tilde{\bm{h}}^{(i)} \sim p_{\bm{\theta}_t}(\bm{h}|c_i)$ as the features for previous task $i < t$ encoded by the current model $\bm{\theta}_t$, and $\hat{\bm{h}}^{(i)} \sim p_{\bm{\theta}_{t-1}}(\bm{h}|c_i)$ as the features for the same task $i$ by the snapshot model $\bm{\theta}_{t-1}$. The first way to construct the feature space is through an adversarially learned encoder $\textit{Enc}$ (Figure~\ref{fig_model}, \textcircled{1} Learned encoder), and the features can be obtained by:
\begin{equation} \label{eq:feature_construction}
\begin{split}
\Tilde{\bm{h}}^{(i)} &= \textit{Enc}(\Tilde{\bm{x}}^{(i)}), \Tilde{\bm{x}}^{(i)} \sim p_{\bm{\theta}_t}(\bm{x}|c_i)\\
\hat{\bm{h}}^{(i)} &= \textit{Enc}(\hat{\bm{x}}^{(i)}), \hat{\bm{x}}^{(i)} \sim p_{\bm{\theta}_{t-1}}(\bm{x}|c_i).
\end{split}
\end{equation}
The \textit{Enc} is trained to compete with a discriminator on distinguishing pairs of $(\Tilde{\bm{h}}, \hat{\bm{h}})$ and $(\hat{\bm{h}}, \hat{\bm{h}})$.

Another way to construct the feature space is by means of knowledge distillation on the intermediate features from the image discriminator (Figure~\ref{fig_model}, \textcircled{2} Distilled knowledge~\footnote{The idea of knowledge distillation for neural networks was proposed by Hinton~\etal~\cite{hinton2015distilling} and was extended to continual learning by Li \etal~\cite{li2018learning} for supervised models.}). This is also similar to the knowledge distillation adopted in Lifelong GAN~\cite{zhai2019lifelong}, where multiple levels of knowledge distillation is used. Alternatively, depending on the availability of prior knowledge on the tasks, we can directly use them as features in the matching process. In this work, we consider features that are extracted from a pretrained model (pretrained on task-irrelevant data) as the representatives of our prior knowledge (Figure~\ref{fig_model}, \textcircled{3} Prior knowledge), due to the unavailability of such information in current benchmark datasets. Note that the use of prior knowledge or task descriptive information does not violate the assumption of continual learning in most use cases, since the true data distributions for previous/future tasks still remain without access. The intuition is that, if a model is already trained to draw $\bigtriangleup$ and $\square$ separately in previous tasks, it should know how to draw $\bigtriangleup \square$ in zero-shot, if we learn a good feature space that renders $\bm{h}_{\bigtriangleup \square} = \bm{h}_{\bigtriangleup} + \bm{h}_{\square}$.
% $E(\bigtriangleup \square) = E(\bigtriangleup) + E(\square)$
This feature space could also be used for our proposed feature matching process, however, the features need to cover the variability of information with respect to the task space.

\subsection{Combined regularization spaces}
In addition to standalone usages, FoCL can also be extended with other regularization strategies depending on the use case scenarios. For example, if the images have high variability in the background, the performance could be further boosted by augmenting FoCL with additional regularization in the image space. In such cases, the objective function can be given as:
\begin{equation}
\begin{split}
  \mathcal{L}_{\alpha\text{-FoCL}}(\bm{\theta}_t) 
  &= \blue{\underbrace{\black{D\infdiv{p_{data}(\bm{x}^{(t)})}{p_{\bm{\theta}_t}(\bm{x} | c_t)} }}_{\text{current task}}} + \blue{\underbrace{\black{ \sum_{i=1}^{t-1}\lambda_t (\alpha D'_{\text{feature}} + (1-\alpha) D'_{\text{image}} )}}_{\parbox{62pt}{\scriptsize\centering previous tasks}}},
  \label{eq:combined_objective}
\end{split}
\end{equation}
% \begin{equation}
% \begin{multlined}
%   \mathcal{L}_{\text{}}(\bm{\theta}_t) = \alpha \mathcal{L}_{\text{feature}}(\bm{\theta}_t) + (1- \alpha) \mathcal{L}_{\text{image}}(\bm{\theta}_t),
%   \label{eq:combined_objective}
% \end{multlined}
% \end{equation}
where $D'_{\text{feature}}=D'\infdiv{p_{\bm{\theta}_{t-1}}(\bm{h}|c_i)}{p_{\bm{\theta}_t}(\bm{h}|c_i)}$, $D'_{\text{image}}=D'\infdiv{p_{\bm{\theta}_{t-1}}(\bm{x}|c_i)}{p_{\bm{\theta}_t}(\bm{x}|c_i)}$, and the choice of the hyperparameter $\alpha$ ($\alpha > 0$) relies on the given data. To distinguish from standalone FoCL, we name this extended version of FoCL as $\alpha$-FoCL.

\subsection{Forgetfulness measurement} \label{sec_fs_method}
The current evaluation metrics used in continual learning research for generative models mostly focus on the overall quality of the generated images for all tasks, for example, the average classification accuracy based on a classifier pretrained the real data (generated images of better quality give better accuracy)~\cite{wu2018memory,ostapenko2019learning,van2018generative}, Fr\'echet inception distance (FID)~\footnote{FID was originally proposed in \cite{heusel2017gans}.}~\cite{wu2018memory,lesort2019generative} and test log-likelihood of the generated samples~\cite{nguyen2017variational}. However, all these metrics fail to disentangle the pure forgetfulness measurement from the generative model performance, \ie, there are two varying factors in current evaluation metrics: the approach to solve the catastrophic forgetting and the choice of using different generative models or architectures (\eg, GAN or VAE). Therefore, we propose a metric that we call \textit{forgetfulness score} for a fair comparison across different methods in continual learning.
% Let $d_t^{(i)}$ be the computed disparity between $p_{\bm{\theta}_t}(\bm{x}^{(t)})$ (\ie, the approximated distribution of task $i$ at sequence $t$) and $p_{data}(\bm{X}^{(t)})$. Also let $J_{ti}=(d_t^{(i)} - d_i^{(i)})$ be the divergence at 
%The forgetness score for taks $i$ measures distance d

\begin{figure}[t]
	\centering
	\includegraphics[width=\linewidth]{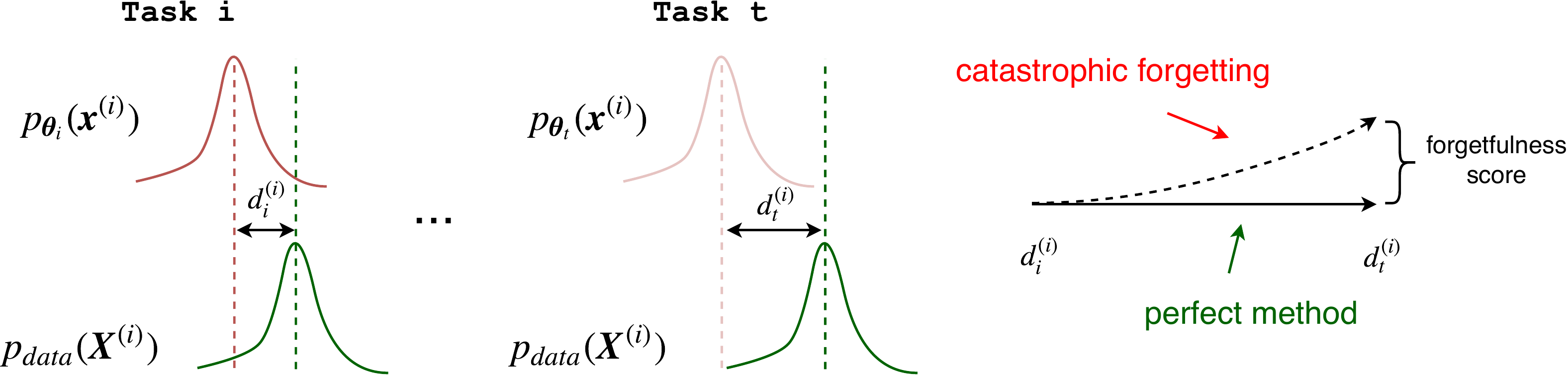}
	\caption{Proposed forgetfulness measurement.}
	\label{fig_eval}
\end{figure}

As shown in Figure \ref{fig_eval}, for each previous task $i \in [1,t-1]$, we compute the distance $d_i^{(i)}$ between the generated data distribution $p_{\bm{\theta}_i}(\bm{x}^{(i)})$ and the true data distribution $p_{data}(\bm{x}^{(i)})$, and at the current task $t$, we recompute the same distance $d_{t}^{(i)}$ (the subscript denotes for the current task index) by using the current model parameterized by $\bm{\theta}_{t}$. The difference ($d_t^{(i)} - d_i^{(i)}$) %between $d_t^{(t)}$ and $d_{t+i}^{(t)}$
measures the amount of forgetting from task $i$ to the current task $t$. % evaluates whether the model suffers from catastrophic forgetting. 
We define $FS_t$ as the task forgetfulness score for task $t > 1$:
\begin{equation}
    % FS = \frac{2}{T*(T-1)} \sum_{t=2}^{T} \sum_{i=1}^{t-1} (d_t^{(i)} - d_i^{(i)}).
    % FS = \frac{1}{T-1} \sum_{t=2}^{T} FS_t,
    % FS_{t} = \frac{1}{t-1} \sum_{i=1}^{t-1} \max(0, (d_t^{(i)} - d_i^{(i)}))
    FS_{t} = \frac{1}{t-1} \sum_{i=1}^{t-1} (d_t^{(i)} - d_i^{(i)}).
  \label{eq:metric}
\end{equation}
For the overall forgetfulness measurement, we average the weighted task forgetfulness scores: $FS = \frac{2}{T*(T-1)} \sum_{t=2}^{T} (t-1)FS_t$. Note that our proposed forgetfulness score requires an assumption that the model is capable of learning the current task well enough so that $d_i^{(i)}$ is meaningful and comparable (\eg, a random model can result in an infinitely large $d_i^{(i)}$, making the subtraction meaningless).
% acts as a complimentary evaluation metric for continual learning methods, since it does not contain information on how well the model learns current tasks. 
% In practice, we find it is unfair to use the original forgetfulness score as a metric when $d_t^{(t)}$ is
In order to compensate this when comparing different methods, we can adjust the original forgetfulness score (eq.~(\ref{eq:metric})) by adding a penalty term on the current task $t$, resulting in compensated forgetfulness score:
\begin{equation}
    CFS_{t} = \frac{1}{t-1} \sum_{i=1}^{t-1} (d_t^{(i)} - d_i^{(i)}) + {d_t^{(t)}}.
  \label{eq:compensated_metric}
\end{equation}
We will show later in the experiments (Section~\ref{sec_fs}) that our forgetfulness measurement offers complementary information that is not available with current commonly used metrics when comparing various methods. In addition, the slope $k$ of the curve for the task forgetfulness score can to a certain degree reveal the potential performance on future tasks.

\section{Experiments}
\subsection{Setup}
\subsubsection{Implementation details}
We follow the similar architecture designs and hyperparameter choices as~\cite{wu2018memory} (code available here \footnote{\url{https://github.com/WuChenshen/MeRGAN}}) in order for a fair comparison. We set $\lambda_t = \frac{1}{t-1}1e\text{-}3$ for both MNIST and Fashion MNIST datasets, and $\lambda_t = \frac{1}{t-1}1e\text{-}2$ for SVHN dataset. For the feature encoder, 3 conv layers are used to extract 128 dimensional vector at $4\times4$ resolution. The feature discriminator is composed of 1 conv layer and 1 linear layer. For the pretrained VGG, we use VGG-19 provided by tensorlayer \footnote{\url{https://tensorlayer.readthedocs.io/en/stable/modules/models.html}}. To reproduce the results from previous methods, we use the implementation from~\cite{nguyen2017variational} for VCL, EWC and SI methods. The code is available here \footnote{\url{ https://github.com/nvcuong/variational-continual-learning}}. For the DGR method, we use the implementation from~\cite{lesort2019generative}, with the code available here \footnote{\url{https://github.com/TLESORT/Generative_Continual_Learning}}. We follow their choices of architecture designs and hyperparameters. %For the SI method,  we set $\lambda$ to 10 instead of $1$, which gives better results in our experiments. 
For the Fashion MNSIT dataset, we train the model for 400 epochs for each task. For computing the average classification accuracies ($A_5$ and $A_{10}$), we evaluate on 6,400 samples by using pretrained classifiers. We use the same classifiers for both MNIST and SVHN datasets provided by~\cite{wu2018memory}, and we train our classifier for Fashion MNIST with ReNet-18. We generate 6,400 images for FID score when calculating our forgetfulness scores.

% We repeat our experiments independently for 3 times when collecting the results for Table~\ref{tab_performance}, Table~\ref{tab_vgg} and Table~\ref{tab_acgan}, and mean $\pm$ std are presented together with the best results obtained. Figure~\ref{fig_iters}, Figure~\ref{fig_afs} and Figure~\ref{fig_eval} use the 68\% confidence interval. All experiments are performed on GeForce GTX 1080 Ti GPUs.

\subsubsection{Datasets}
We perform our experiments on four benchmark datasets: MNIST~\cite{lecun1998mnist}, SVHN~\cite{netzer2011reading}, Fashion MNIST~\cite{xiao2017fashion} and CIFAR10~\cite{krizhevsky2009learning}. SVHN contains 73,257 training and 26,032 test 32$\times$32 pixel color images of house number digits. MNIST and Fashion MNIST contain ten categories of black-and-white 28$\times$28 pixel images composed of 60,000 training and 10,000 test images. Images from these two datasets are zero-padded the borders to form 32$\times$32 images. 
% These Both of these  and resized the images from 28$\times$28 to 32$\times$32 by zero-padding the borders. The datasets are composed of ten categories of 60,000 training images and 10,000 test-images
% SVHN, 10 classes, black and white Vs. RGB, 
% \cite{nguyen2017variational}
CIFAR10 has 50,000 training images and 10,000 test images in 10 classes, and all images are 32$\times$32 color images.
% To train the models continually, each dataset was broken into 10 sequential tasks.
Following previous works~\cite{wu2018memory, ostapenko2019learning}, we also consider each class as a different task in this work.

\subsection{Adversarially learned feature encoder} \label{exp_end_to_end}
In this subsection, we construct our feature space by adversarially learning the matched and mismatched pairs of features. The features are extracted from an encoder that plays a minmax game with a discriminator. We use the WGAN-GP~\cite{gulrajani2017improved} technique in training our encoder, which makes our divergence function the Wasserstein distance when matching the empirical feature distributions: $W(p_{\bm{\theta}_{t-1}}(\bm{h}^{(i)}), p_{\bm{\theta}_{t}}(\bm{h}^{(i)}))$ for all $i < t$ at task $t$. Similar to previous work~\cite{wu2018memory}, we use conditional batch normalization~\cite{de2017modulating} for the task conditioning, and integrate an auxiliary classifier (AC-GAN)~\cite{odena2017conditional} to predict the category labels. The weight of auxiliary classifier for AC-GAN is set to $1$. %We perform our experiments on four benchmark datasets: MNIST~\cite{lecun1998mnist}, SVHN~\cite{netzer2011reading}, Fashion MNIST~\cite{xiao2017fashion} and CIFAR10~\cite{krizhevsky2009learning}. More details on implementation and datasets are provided in the supplementary material.

In order to compare our proposed FoCL (standalone) with previous methods, we first use the same quantitative metric (average classification accuracy) as used in previous works~\cite{wu2018memory,ostapenko2019learning} for both 5 sequential tasks ($A_5$) and 10 sequential tasks ($A_{10}$). As shown in Table~\ref{tab_performance}, FoCL achieves better performance for 10 tasks ($A_{10}$) on three datasets. On the Fashion MNIST dataset, it remarkably improves the accuracy from 80.46\% to 90.26\%. For 5 tasks ($A_5$), FoCL also outperforms previous state-of-the-art method MeRGAN~\cite{wu2018memory} on the MNIST and SVHN datasets. However, it has an impaired result for $A_5$ on Fashion MNIST dataset. Generally, we observe %a consistent phenomenon 
that FoCL tends to consistently work better as the number of tasks grows, which is within our expectation since constructing a representative feature space typically requires training on a certain amount of samples and tasks. It could be possible that our feature encoder has not been trained well in early tasks, and this agrees with our results in Section~\ref{sec_prior_knowledge}, where $A_5$ is shown to be improved by using prior knowledge as feature encoder ( \ie, the feature space is pre-constructed).

\begin{table*}[t]
    \caption{Performance comparison based on average classification accuracy.}
    \begin{adjustbox}{width=\textwidth}
      \begin{tabular}{r*{7}{c}}
        \toprule
        & \multicolumn{2}{c}{MNIST (\%)} & \multicolumn{2}{c}{SVHN (\%)} & \multicolumn{2}{c}{Fashion MNIST (\%)} \\
        \cmidrule(lr){2-3} \cmidrule(lr){4-5} \cmidrule(lr){6-7}
        Method & $A_5$ & $A_{10}$ & $A_5$ & $A_{10}$ & $A_5$ & $A_{10}$ \\
        \midrule
        JT & 97.66 & 96.92 & 85.30 & 84.82 & 87.12 & 89.08 \\
        \midrule
        EWC \cite{seff2017continual} & 70.62 & 77.03 & 39.84 & 33.02 & - & - \\
        DGR \cite{shin2017continual} & 90.39 & 85.40 & 61.29  & 47.28 & - & - \\
        MeRGAN \cite{wu2018memory} & 98.19 & 97.01 & 80.90 & 66.78 & \textbf{92.17}* & 80.46* \\
        % DGM \cite{ostapenko2019learning} & \textbf{99.17} & 97.92 & 83.93 & 74.38 & - & - \\
        \midrule
        FoCL $\dagger$ & \textbf{99.07} & \textbf{98.09} & \textbf{84.80} & \textbf{77.31} & 86.18 & \textbf{90.26} \\
        (mean $\pm$ std) & 98.96 $\pm$ 0.13 & 97.68 $\pm$ 0.25 & 84.14 $\pm$ 0.58 & 76.49 $\pm$ 0.65 & 85.02 $\pm$ 0.82 & 89.58 $\pm$ 0.40 \\
        \bottomrule
		\multicolumn{7}{l}{\footnotesize * results based on our experiments; $\dagger$ standalone FoCL, best result based on 3 independent experiments. }
      \end{tabular}
    \end{adjustbox}
    \label{tab_performance}
\end{table*}
\begin{figure*}[!t]
	\centering
	\includegraphics[width=\textwidth]{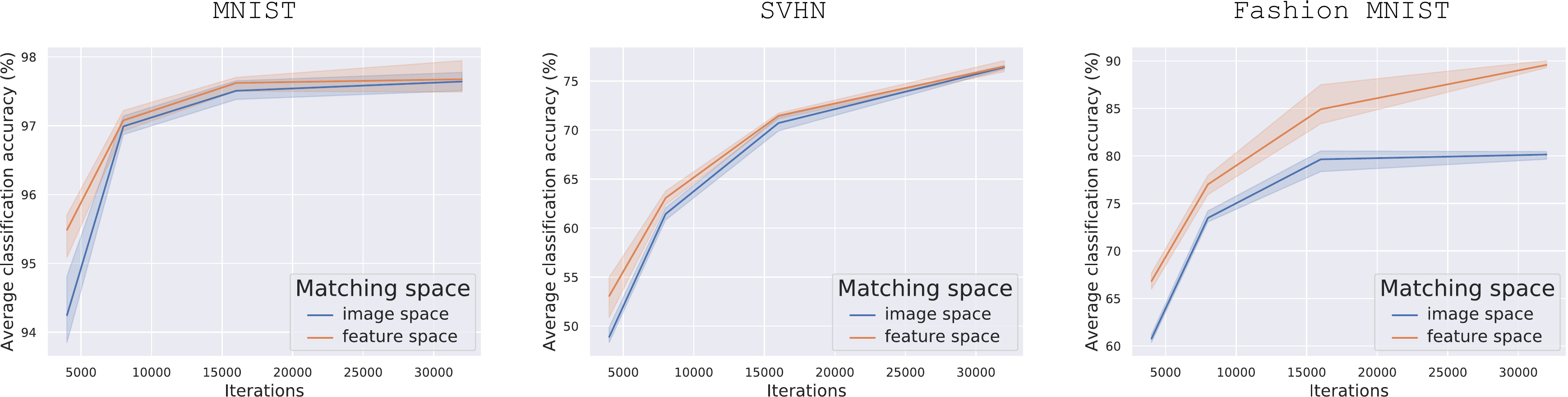}
	\caption{Fast convergence in feature space for 10 tasks.}
	\label{fig_iters}
\end{figure*}

\begin{figure*}[htbp]
	\centering
	\includegraphics[width=\linewidth]{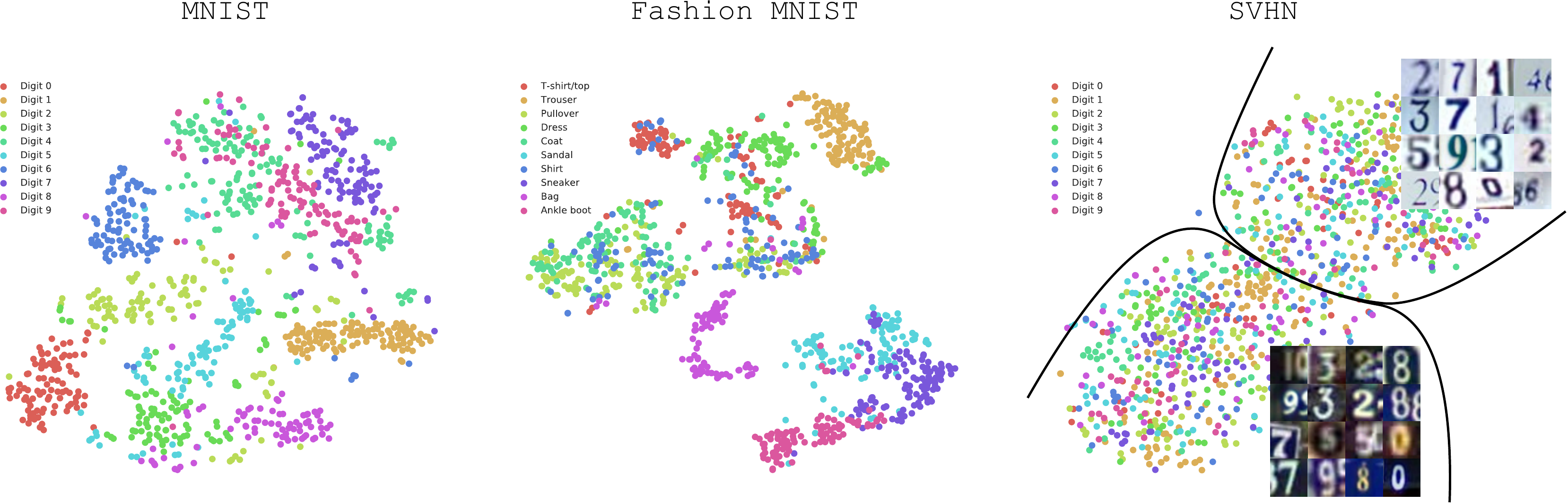}
	\caption{t-SNE visualizations of encoded features (extracted from adversarially learned encoders).}
	\label{fig_features}
\end{figure*}

We also find in our experiments that, the performance can be improved by simply increasing the number of iteration steps when using the generative replay method~\cite{shin2017continual}. We then further investigate how replay iteration affects the model performance when we switch the regularization from image space to feature space. Figure~\ref{fig_iters} demonstrates that not only can FoCL give significantly better performance (\eg, Fashion MNIST), but it can also have faster adaptation, suggesting the efficiency of matching in the feature space as compared to the image space where many pixels can be perturbation factors. This observation also agrees with the findings in~\cite{bengio2019meta} that high-level representation space can lead to fast adaptation to distributional changes due to non-stationarities.

\begin{table*}[!t]
    \caption{Auxiliary classifier (AC) in the feature encoder on SVHN dataset.}
    \begin{adjustbox}{width=\textwidth}
      \begin{tabular}{{l}c*{7}{c}}
        \toprule
        Weight & \multicolumn{8}{c}{Iterations per task (\%)} \\
        \cmidrule(lr){2-9}
        on & \multicolumn{2}{c}{4K} & \multicolumn{2}{c}{8K} & \multicolumn{2}{c}{16K} & \multicolumn{2}{c}{32K} \\
        \cmidrule(lr){2-3} \cmidrule(lr){4-5} \cmidrule(lr){6-7} \cmidrule(lr){8-9}
        AC & $A_5$ & $A_{10}$ & $A_5$ & $A_{10}$ & $A_5$ & $A_{10}$ & $A_5$ & $A_{10}$ \\
        \midrule
        $\lambda$=0 * & 62.26 $\pm$ 0.99 & 53.06 $\pm$ 1.69 & 71.90 $\pm$ 0.53 & 63.07 $\pm$ 0.77 & 80.18 $\pm$ 0.73 & 71.47 $\pm$ 0.31 & 84.14 $\pm$ 0.58 & 76.49 $\pm$ 0.65 \\
        $\lambda$=1 & 61.33 $\pm$ 1.28 & 52.50 $\pm$ 0.54 & 72.55 $\pm$ 0.73 & 63.17 $\pm$ 0.15 & 79.58 $\pm$ 1.09 & 71.56 $\pm$ 0.70 & 84.23 $\pm$ 0.58 & 76.72 $\pm$ 0.75 \\
        $\lambda$=1e\text{-}3 & 47.13 $\pm$ 1.26 & 36.50 $\pm$ 1.64 & 65.35 $\pm$ 2.33 & 53.88 $\pm$ 0.47 & 75.85 $\pm$ 0.58 & 64.33 $\pm$ 2.38 & 84.44 $\pm$ 0.15 & 75.93 $\pm$ 0.32 \\
        $\lambda$=1e\text{-}5 & 45.20 $\pm$ 2.89 & 35.57 $\pm$ 1.92 & 66.61 $\pm$ 0.33 & 56.03 $\pm$ 0.53 & 75.71 $\pm$ 1.07 & 66.22 $\pm$ 0.88 & 82.49 $\pm$ 0.30 & 72.21 $\pm$ 0.36 \\
        \bottomrule
        \multicolumn{7}{l}{\footnotesize * without AC in the feature encoder }
      \end{tabular}
    \end{adjustbox}
    \label{tab_acgan}
\end{table*}

To visualize the learned features, we encode the testing images in the datasets for t-SNE visualization (Figure~\ref{fig_features}). In both MNIST and Fashion MNIST datasets, our learned features are task-discriminative whereas in SVHN dataset, the features contain task agnostic information such as background color. The correlation between features being task-discriminative and the better performance in classification accuracy raises a question whether the former is a causal explanation of the latter. To answer this question, we integrate an auxiliary classifier in our encoder to further regularize the features for SVHN dataset to be task-discriminative. However, we do not observe significant performance changes (Table~\ref{tab_acgan}). Interestingly, we find that in both MNIST and Fashion MNIST datasets, the embeddings for the last task (\ie, \textit{digit 9} and \textit{ankle boot}) have been meaningfully allocated in the feature space (\textit{digit 9} close to \textit{digit 4} and \textit{digit 7}, and \textit{ankle boot} close to \textit{sandal} and \textit{sneaker}), even though the images of the last task have not been exposed to the encoder (Note that the encoder is trained during the replay process only on previous tasks). This implies an important future direction in continual learning to incrementally learn task descriptive features that can be potentially applied to zero-shot learning scenarios.

\begin{table*}[t]
    \caption{Different ways to construct the feature space.}
    \centering
    \begin{adjustbox}{width=\textwidth}
      \begin{tabular}{r*{7}{c}}
        \toprule
        & \multicolumn{2}{c}{MNIST (\%)} & \multicolumn{2}{c}{SVHN (\%)} & \multicolumn{2}{c}{Fashion MNIST (\%)} \\
        \cmidrule(lr){2-3} \cmidrule(lr){4-5} \cmidrule(lr){6-7}
        Feature source & $A_5$ & $A_{10}$ & $A_5$ & $A_{10}$ & $A_5$ & $A_{10}$ \\
        % \midrule
        % JT & 97.66 & 96.92 & 85.30 & 84.82 & ? & ? \\
        \midrule
        % Ours & 99.07 & \textbf{98.09} & \textbf{84.80} & \textbf{77.31} & 86.18 & \textbf{90.26} \\
        Learned encoder & \textbf{98.96 $\pm$ 0.13} & \textbf{97.68 $\pm$ 0.25} & 84.14 $\pm$ 0.58 & 76.49 $\pm$ 0.65 & 85.02 $\pm$ 0.82 & \textbf{89.58 $\pm$ 0.40} \\
        % vgg features & 24.75 & 15.32 & 85.09 & 54.51 & 93.61 & 65.31 \\
        Distilled knowledge & 98.67 $\pm$ 0.10 & {\bf 97.57 $\pm$ 0.15}& 83.67 $\pm$ 2.54 & {\bf 77.63 $\pm$ 0.10} &	\textbf{90.18 $\pm$ 0.19} &	{\bf 89.40 $\pm$ 0.35}\\
        Prior knowledge & 21.53 $\pm$ 2.28 & 13.15 $\pm$ 1.72 & \textbf{85.02 $\pm$ 0.08} & 54.46 $\pm$ 0.06 & \textbf{90.39 $\pm$ 4.01} & 57.21 $\pm$ 6.79 \\
        % vgg features & 21.53 $\pm$ 2.28 & 13.15 $\pm$ 1.72 & 85.02 $\pm$ 0.08 & 54.46 $\pm$ 0.06 & 93.21 $\pm$ 0.39 & 61.47 $\pm$ 3.84 \\
        \bottomrule
      \end{tabular}
    \end{adjustbox}
    \label{tab_vgg}
\end{table*}

\subsection{Distilled or prior knowledge as feature encoder} \label{sec_prior_knowledge}
Next, we consider features either achieved from an intermediate layer of the image discriminator as distilled knowledge, or extracted from a pretrained model (\eg, VGG pretrained on ImageNet) as prior knowledge for feature matching during the replay process. Since the features are not adversarially learned in both cases, we simply use $l_2$ loss to align the features, therefore the regularization is equivalent to the use of Bregman divergence with $\phi(x) = \left\|x\right\|^2$ for the divergence function $D'$ in eq.~(\ref{eq:feature_objective}). Table~\ref{tab_vgg} shows the performance of FoCL with different feature sources. Surprisingly, using prior knowledge as feature source gives improved performance in $A_5$ on both SVHN (85.09\%, best result) and Fashion MNIST (93.61\%, best result) datasets, 
%, even though the encoded features may not be perfect to represent the true data. The best $A_5$ is 85.09\% on SVHN dataset and 93.61\% on Fashion MNIST dataset, 
which also outperform the results obtained from using regularization in the image space (Table~\ref{tab_performance}, MeRGAN~\cite{wu2018memory}). However, this finding is neither generalizable (MNIST) nor scalable ($A_{10}$). In most cases, we obtain comparable performance between the adversarially learned encoder and distilled knowledge as feature sources, suggesting that the improved performance comes from the regularization in the feature space rather than the choice of divergence function.

\subsection{Boosted performance with $\alpha$-FoCL}
\begin{figure*}[t]
	\centering
	\includegraphics[width=\textwidth]{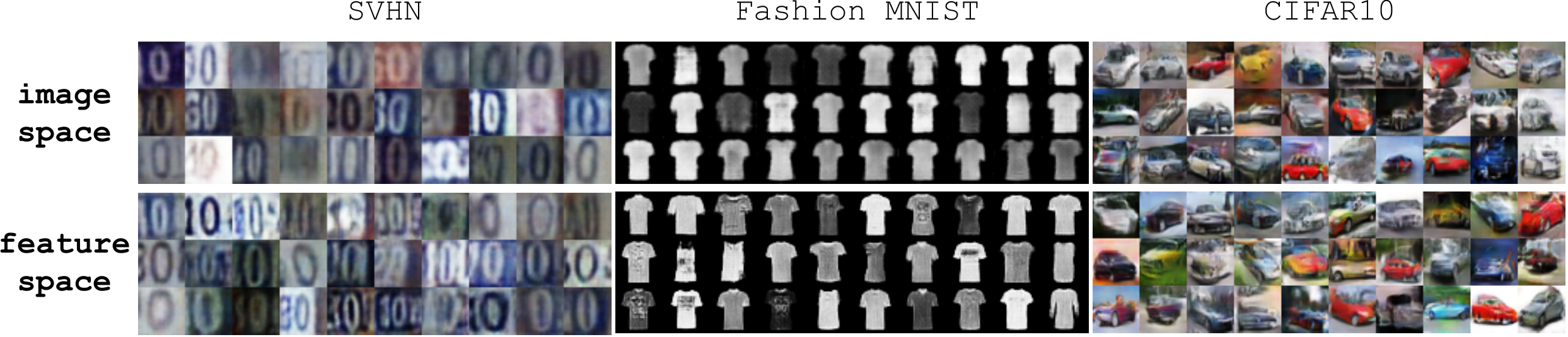}
	\caption{Comparison of generated samples between regularization in image space and feature space.}
	\label{fig_images}
\end{figure*}

Despite the improved performance with standalone FoCL compared to previous methods, we notice in our experiments that the regularization in the feature space alone is insufficient to address catastrophic forgetting when dealing with image data that has high variability in the background, such as the CIFAR10 dataset (Table~\ref{tab_alpha_all}), possibly due to the less attention on fine-grained background details in our proposed approach. However, this can be effectively addressed by $\alpha$-FoCL (eq.~(\ref{eq:combined_objective})). Moreover, with $\alpha$-FoCL, we also observe additionally boosted performance on MNIST, SVHN and Fashion MNIST datasets compared to standalone FoCL, achieving the new state-of-the-art results (Table~\ref{tab_alpha_all}). Figure~\ref{fig_images} compares the generated samples between image space and feature space, and the latter has notable improvement on sharpness (SVHN) and diversity (Fashion MNIST).

\begin{table}[!t]
    \caption{Boosted performance with $\alpha$-FoCL.}
    \centering
    \begin{adjustbox}{width=\linewidth}
      \begin{tabular}{{lcccc}}
        \toprule
        Method & MNIST(\%) & SVHN(\%) & F-MNIST(\%) & CIFAR10(\%) \\
        \midrule
        % DGM \cite{ostapenko2019learning} & 97.92 & 74.38 & - & 56.21 \\
        MeRGAN~\cite{wu2018memory}* & 97.81 & 76.68 & 80.46 & 68.92 \\
        \midrule
        {$\alpha$=0.0} (equivalent to MeRGAN) & 97.81 & 76.68 & 80.46 & 68.92 \\
        {$\alpha$=0.2} & 97.72 & 78.13 & 87.11 & 70.65 \\        
        {$\alpha$=0.4} & 98.01 & \textbf{78.95} & 90.63 & 71.29 \\
        {$\alpha$=0.6} & 97.94 & 77.70 & 90.32 & 71.35 \\
        {$\alpha$=0.8} & 97.89 & 77.59 & \textbf{91.73} & \textbf{71.58} \\
        {$\alpha$=1.0} (equivalent to standalone FoCL) & \textbf{98.09} & 77.31 & 90.26 & 23.85 \\
        \bottomrule
        \multicolumn{5}{l}{\footnotesize * improved results based on our experiments} \\
        % \multicolumn{5}{l}{\footnotesize $\dagger$ equivalent to MeRGAN~\cite{wu2018memory}} \\
        % \multicolumn{5}{l}{\footnotesize $\ddagger$ equivalent to standalone FoCL}
      \end{tabular}
    \end{adjustbox}
    \label{tab_alpha_all}
\end{table}

\subsection{Forgetfulness evaluation} \label{sec_fs}
Here, we further evaluate FoCL (standalone) using our proposed forgetfulness scores. We choose to compute our task forgetfulness scores by using FID as the distance measure between the generated data and true data \footnote{Another choice for distance measure could be based on classification accuracy.}. As discussed before, in practice, we also notice that the original forgetfulness score ($FS$, eq.~(\ref{eq:metric})) is only meaningful when $d_i^{(i)}$ is comparable, whereas the compensated forgetfulness score ($CFS$, eq.~(\ref{eq:compensated_metric})) is more suitable when comparing different methods with big variations in $d_i^{(i)}$, as the latter also takes into consideration the difficulty level of not forgetting. %Heuristically, it is more difficult for a model that has learned all fine-grained details (small $d_t^{(t)}$) to keep remembering all the information, compared to a random model (big $d_t^{(t)}$). 

\begin{figure}[!t]
	\centering
	\includegraphics[width=\textwidth]{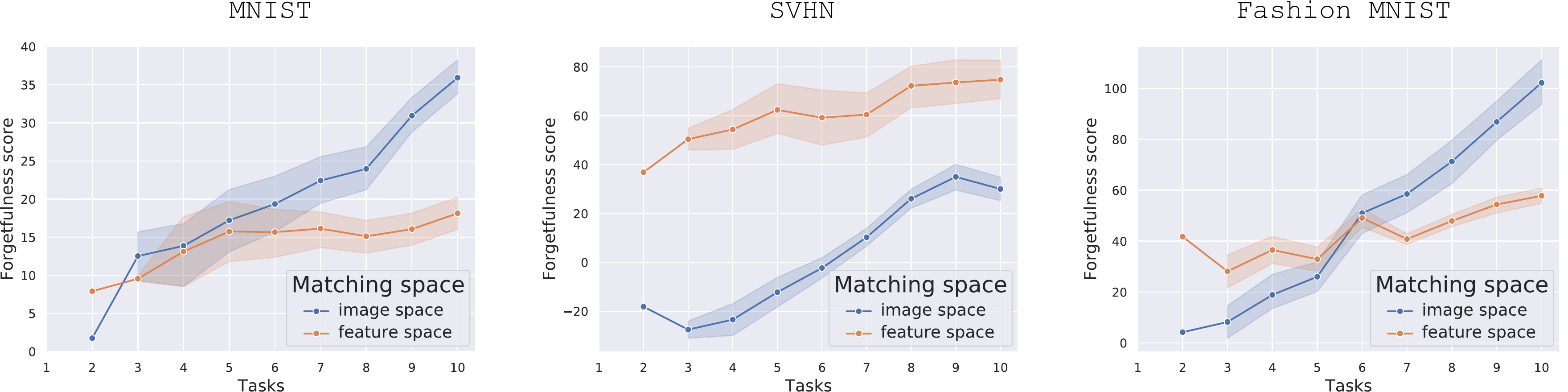}
	\caption{Forgetfulness score by tasks ($FS \downarrow$).}
	\label{fig_fs}
\end{figure}

\begin{table}[!t]
    \caption{Overall forgetfulness score ($FS \downarrow$) based on FID.}
    \centering
    % \begin{adjustbox}{width=\textwidth}
      \begin{tabular}{r*{10}{c}}
        \toprule
        %  & \multicolumn{3}{c}{FID based} & \multicolumn{3}{c}{Classification accuracy based (FS/slope)}\\
        % \cmidrule(lr){2-4} \cmidrule(lr){5-7}
         & \multicolumn{2}{c}{MNIST} & \multicolumn{2}{c}{SVHN} & \multicolumn{2}{c}{Fashion MNIST} \\
        \cmidrule(lr){2-3} \cmidrule(lr){4-5} \cmidrule(lr){6-7}
        Matching space & $FS \downarrow$ & $k \downarrow$ & $FS \downarrow$ & $k \downarrow$ & $FS \downarrow$ & $k \downarrow$ \\
        \midrule
        % EWC \cite{seff2017continual} \\
        % DGR \cite{shin2017continual} & 32.57 & 1.53 & & & \textbf{34.73} & 4.19 \\
        % SI \cite{zenke2017continual} \\
        % VCL \cite{nguyen2017variational} \\
        % not weighted
        % MeRGAN \cite{wu2018memory} & 19.77 & 3.62 & 11.26 & 9.04 & 47.48 & 13.46 \\
        % ours & 14.16 & 0.82 & 60.52 & 3.81 & 43.27 & 3.59 \\
        Image space & 24.60 & 3.62 & \textbf{13.12} & 9.04 & 64.48 & 13.46 \\
        Feature space & \textbf{15.60} & \textbf{0.82} & 66.19 & \textbf{3.81} & \textbf{47.13} & \textbf{3.59} \\
        \bottomrule
      \end{tabular}
    % \end{adjustbox}
    \label{tab_fs}
\end{table}

\begin{figure*}[t]
	\centering
	\includegraphics[width=\textwidth]{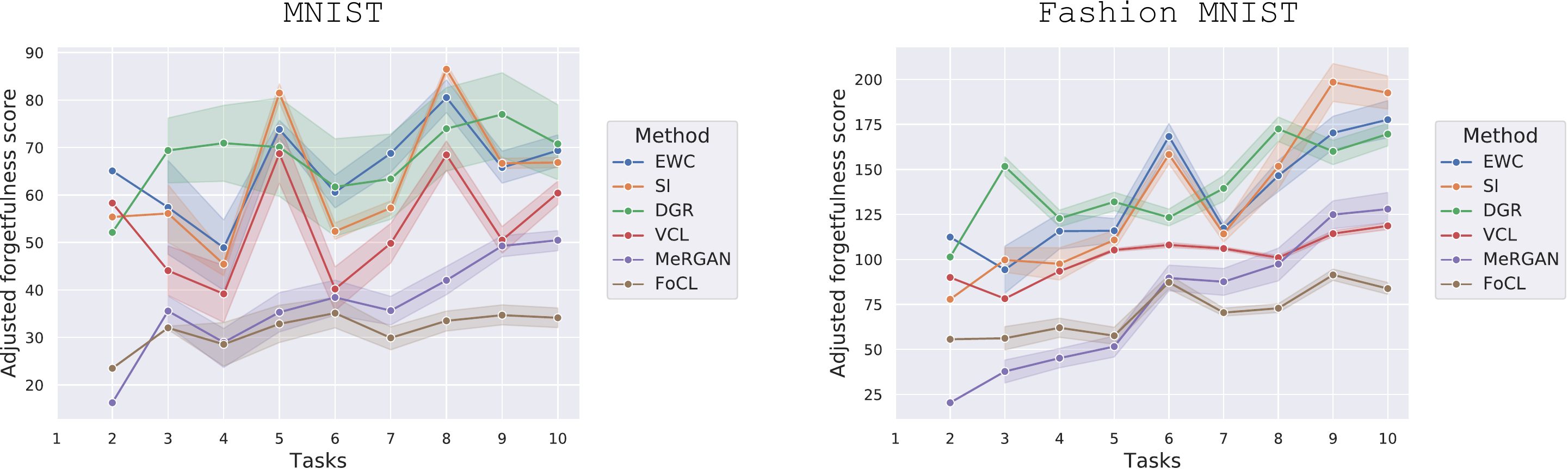}
	\caption{The curves of compensated forgetfulness score ($CFS \downarrow$).}
	\label{fig_afs}
\end{figure*}

Therefore, in our experiments, we %use $CFS$ for comparison among different methods and $FS$ for 
use $FS$ only for the comparison between the regularization in the image space and feature space, where the experiments are more strictly controlled so that the assumption of $d_i^{(i)}$ being comparable is satisfied. The results of $FS$ are shown in Figure~\ref{fig_fs}, and Figure~\ref{fig_afs} compares the curve of $CFS$ from FoCL to those from previous methods including DGR~\cite{shin2017continual} and MeRGAN~\cite{wu2018memory} that use the regularization in the image space. %, and Figure~\ref{fig_fs} (supplementary material) shows the comparison of the original task forgetfulness score $FS$ between the regularization in the image space and feature space. 
As seen in both figures (more evident in MINST and Fashion MNIST datasets for $FS$), despite the latency at early tasks compared to MeRGAN, FoCL can quickly catch up in performance and maintain a more stable state of forgetfulness, suggesting that it suffers less from the forgetting as the task goes on. The previous methods, however, have much steeper curves. On SVHN dataset (Figure~\ref{fig_fs}), we do not observe lower $FS$ from FoCL possibly due to the limited number of tasks in SVHN given the complexity of the data, which has not allowed FoCL to outperform. %Another possible explanation could be the high FID values (>100) for both methods that make the distance measure not reliable for computing our forgetfulness score. 

\begin{table}[!t]
    \caption{Comparison of different methods by the overall compensated forgetfulness score ($CFS \downarrow$) based on FID.}
    \centering
    % \begin{adjustbox}{width=\textwidth}
      \begin{tabular}{r*{5}{c}}
        \toprule
        & \multicolumn{2}{c}{MNIST (\%)} & \multicolumn{2}{c}{Fashion MNIST (\%)} \\
        \cmidrule(lr){2-3} \cmidrule(lr){4-5}
        Method & $CFS \downarrow$ & $k \downarrow$ & $CFS \downarrow$ & $k \downarrow$ \\
        \midrule
        EWC \cite{seff2017continual} & 67.82 & 1.61 & 147.63 & 10.02 \\
        SI \cite{zenke2017continual} & 66.13 & 1.99 & 152.70 & 15.46 \\
        DGR \cite{shin2017continual} & 69.86 & 1.31 & 150.39 & 7.24 \\
        VCL \cite{nguyen2017variational} & 54.79 & 1.59 & 106.94 & \textbf{3.80} \\
        MeRGAN \cite{wu2018memory} & 41.43 & 3.29 & 94.30 & 13.86 \\
        \midrule
        FoCL & \textbf{32.87} & \textbf{0.68} & \textbf{76.41} & 4.12 \\
        \bottomrule
      \end{tabular}
    % \end{adjustbox}
    \label{tab_afs}
\end{table}

Another possible explanation could be that our method focuses more on task representative information rather than all fine-grained details, therefore resulting in worse FID scores, however, it manages to remember more task representative information and as a result gives better classification accuracy (Table~\ref{tab_performance}). Nevertheless, in general, we still observe that the growth of forgetfulness score is slowed down when the matching space is in the feature space instead of image space. In fact, the trend of the curve can also offer us a prediction on the forgetfulness score for future tasks, and we can use the slope $k$ of the linearly-fitted curve as the indicator. Table~\ref{tab_fs} and Table~\ref{tab_afs} summarize the comparisons of both the overall forgetfulness scores and their corresponding linearly-fitted curve slope $k$ for different methods.

Our forgetfulness measurement gives complementary information in addition to previous metrics such as the average classification accuracy shown in Table~\ref{tab_performance}. %, when evaluating different methods for continual learning. 
For example, we show on MNIST dataset that FoCL has significantly better performance in forgetfulness scores (Table~\ref{tab_fs} and Table~\ref{tab_afs}), while it is indistinguishable in average classification accuracy (Table~\ref{tab_performance}). In fact, the measure of forgetfulness scores focuses on the degree to which the model suffers from catastrophic forgetting during the learning process while the previous metrics merely care about whether the end point can successfully solve all the tasks. We believe both of them should be taken into consideration in order for a fair comparison among different methods in continual learning for future studies.

\section{Conclusion}
In this paper, we present FoCL, a general framework in continual learning for generative models. Instead of the regularization in the parameter space or image space as done in previous works, FoCL regularizes the model in the feature space. To construct the feature space, several ways have been proposed. We show in our experiments on several benchmark datasets that FoCL has faster adaptation and achieves the state-of-the-art performance for generative models in task incremental learning. Finally, we show our proposed forgetfulness measurement offers a new perspective view when evaluating different methods in continual learning.

\bibliography{elsarticle-template}

\end{document}